\documentclass[10pt, conference, compsocconf]{IEEEtran}

\usepackage{cite}
\ifCLASSINFOpdf
\else
\fi

\usepackage{graphicx}
\graphicspath{{images/}}
\usepackage{float}
\usepackage{subcaption}
\usepackage{algpseudocode}
\usepackage{algorithm}
\usepackage{hyperref}

\usepackage{array, bm}
\usepackage{multirow}
\usepackage{twoopt}
\usepackage{tikz}
\usepackage{xcolor}

\usetikzlibrary{fit}
\newcommand\tikzmark[1]{%
  \tikz[remember picture,overlay]\node[inner xsep=0pt] (#1) {};}
\newcommandtwoopt\Textbox[5][5cm][2cm]{\begin{tikzpicture}[remember picture,overlay] \coordinate (aux) at ([xshift=#1]#4); \node[inner ysep=3pt,yshift=0.6ex,draw=red,thick, fit=(#3) (aux),baseline] (box) {}; \node[text width=#2,anchor=north east, font=\sffamily\footnotesize,align=right] at (box.north east) {#5}; \end{tikzpicture}}
\newcommandtwoopt\TextboxSmall[5][1cm][2cm]{\begin{tikzpicture}[remember picture,overlay] \coordinate (aux) at ([xshift=#1]#4); \node[inner ysep=3pt,yshift=0.6ex,draw=red,thick, fit=(#3) (aux),baseline] (box) {}; \node[text width=#2,anchor=north east, font=\sffamily\footnotesize,align=right] at (box.north east) {#5}; \end{tikzpicture}}

\begin{document}

\title{\textbf{\LARGE HyperMODEST: Self-Supervised 3D Object Detection with Confidence Score Filtering}}

\author{\IEEEauthorblockN{Jenny Xu}
\IEEEauthorblockA{\emph{Department of Computer Science}\\
\emph{University of Toronto} \\
\emph{Toronto, Canada}\\
\emph{jennyziyi.xu@mail.utoronto.ca}}
\and
\IEEEauthorblockN{Steven L. Waslander}
\IEEEauthorblockA{\emph{Institute for Aerospace Studies}\\
\emph{University of Toronto} \\
\emph{Toronto, Canada}\\
\emph{steven.waslander@robotics.utias.utoronto.ca}}
}
\maketitle

\begin{abstract}

Current LiDAR-based 3D object detectors for autonomous driving are almost entirely trained on human-annotated data collected in specific geographical domains with specific sensor setups, making it difficult to adapt to a different domain. MODEST \cite{modest} is the first work to train 3D object detectors without any labels. Our work, HyperMODEST, proposes a universal method implemented on top of MODEST that can largely accelerate the self-training process and does not require tuning on a specific dataset. We filter intermediate pseudo-labels used for data augmentation with low confidence scores. On the nuScenes dataset, we observe a significant improvement of 1.6\% in AP\_{BEV} in 0-80m range at IoU=0.25 and an improvement of 1.7\% in AP\_{BEV} in 0-80m range at IoU=0.5 while only using one-fifth of the training time in the original approach by MODEST\cite{modest}. On the Lyft dataset, we also observe an improvement over the baseline during the first round of iterative self-training. We explore the trade-off between high precision and high recall in the early stage of the self-training process by comparing our proposed method with two other score filtering methods: confidence score filtering for pseudo-labels with and without static label retention. The code and models of this work are available at \url{https://github.com/TRAILab/HyperMODEST}	
\end{abstract}


\section{INTRODUCTION}

3D object detection is essential for self-driving cars to navigate safely in highly dynamic scenarios, as vehicles must detect and localize other road agents accurately at all times. With the improvement of LiDAR technologies, LiDAR-based 3D object detectors \cite{centerpoint(2021)}\cite{pv-rcnn++(2022)}\cite{AFDet(2020)} have achieved considerable progress in the last decade. However, the data collection and annotation processes are often performed manually and are time and resource-consuming. The representativeness of the curated data is also always limited in terms of geographical location, weather conditions and sensor modality. For example, \cite{germany} concluded that the variation in car size in Germany and the USA greatly influenced the object detection result. 

\begin{figure}[!t]
	\centering 
	\includegraphics[width=2.9in]{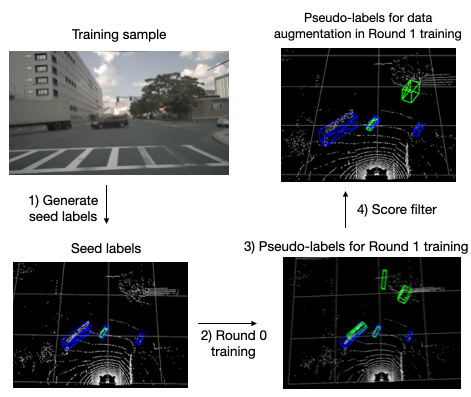} 
	\caption{An illustration of HyperMODEST, our method of score filtering on nuScenes dataset. 1), Bottom left: Seed labels are generated from the unannotated training sample according to MODEST \cite{modest}. 2), Round 0 training with seed labels as pseudo ground-truth. 3), Bottom right: Detections from Round 0 training are used as pseudo-labels for Round 1 training. 4), Only pseudo-labels with high confidence scores are used for data augmentation.}
	\label{fig:intro_nuscenes_1984}
\end{figure}

\begin{figure}[!t]
	\centering 
	\includegraphics[width=2.9in]{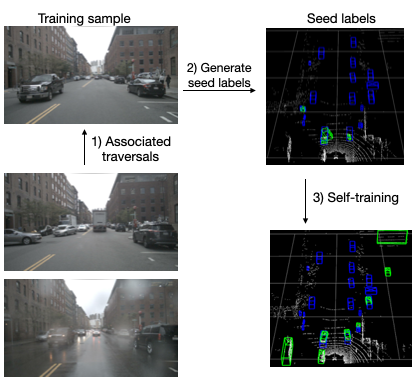} 
	\caption{An illustration of MODEST (Mobile Object Detection with Ephemerality and Self-Training) \cite{modest}. 1), Multiple traversals of the same location are associated with the training sample. 2), Seed labels generation based on point ephemerality. 3), Self-training based on seed labels.  }
	\label{fig:intro_nuscenes_2976}
\end{figure}

MODEST (Mobile Object Detection with Ephemerality and Self-Training)\cite{modest} is the first work to train object detectors without any labels at all. We illustrate its architecture in Fig. \ref{fig:intro_nuscenes_2976}. MODEST includes two stages: seed label generation and self-training. In the first stage, for multiple traversals of the same scene, an ephemerality score that captures the change of the LiDAR point's local neighbourhood across traversals is calculated for each LiDAR point, highlighting moving objects which need to be detected. Then, the LiDAR points are clustered with the DBSCAN \cite{dbscan} clustering algorithm according to the ephemerality scores and their 3D coordinates. The high ephemerality clusters are chosen as the seed labels that represent a useful subset of objects to be detected. In the second stage, a 3D object detector is trained iteratively over multiple rounds of training. In the first round, the model is trained from scratch on the seed labels. The trained model is used to produce an improved set of pseudo ground-truth labels for the second round of training. In the second round, iterative self-training allows the model to produce higher-quality bounding boxes, and the process iterates.

We chose to explore MODEST because of its simple architecture and impressive object detection performance and generalizability. The seed labels generation stage uses a simple but generalizable approach. The seed label generation does not rely on any specific dataset domain. Instead, it relies on the fact that mobile objects are unlikely to stay persistent at the same location over multiple traversals. Self-training allows the model to improve itself over time and discover more objects than the set of seed labels. Similar to seed label generation, self-training is an automatic and generalizable process. Since human drivers usually traverse the same routes daily, this promising self-training technique can be used for edge computing on each customer's autonomous car. 

We present several areas for improvement. Since the seed labels are conservative and high quality, they only capture a small subset of all objects. In addition, the seed labels are filtered based on common heuristics, such as the minimum number of points in the cluster and distance above the ground. Thus, the labels are noisy and contain inaccurate bounding boxes. Moreover, since MODEST is an entirely self-supervised pipeline, self-training can put overconfident label beliefs on noisy data, leading to the propagation of errors across rounds. Supervision on selecting pseudo-labels would be beneficial, but this is infeasible since we enforce the requirement of zero access to ground truth labels during the training process.  

In this paper, we propose HyperMODEST, a generalizable technique of filtering intermediate pseudo-labels used for data augmentation with low confidence scores implemented on top of MODEST, as shown in Fig. \ref{fig:intro_nuscenes_1984}. We also compare this strategy with score filtering on the pseudo-labels with and without static label retention. We are motivated by the following insights. 

First, we leverage curriculum learning \cite{curriculum_1, curriculum_2} to filter out pseudo-labels with low confidence scores for models to learn easier examples. This is a form of indirect supervision. Second, data augmentation \cite{second} is an indispensable technique aimed at increasing the generalizability of a possibly overfitting model. We observe that filtering intermediate pseudo-labels used for data augmentation helps the model outperform the original implementation using only one-fifth of the original training time. We also observe that filtering intermediate pseudo-labels used in data augmentation achieves better performance than filtering intermediate pseudo-labels directly. In the early rounds of training, it is more important to achieve high recall for the pseudo-labels and allow the model to distinguish between true positives and false positives over time. Third, our method of score filtering for labels used in data augmentation is dataset-independent. We believe that this technique can accelerate self-training and help the model to achieve higher performance without the need for extensive tuning. In this technique, we still preserve all outputs from the previous round of training in the pseudo-labels for the current round of training. Thus, we don't need to worry about losing true positives. 

Our contributions are three-fold. 

\begin{enumerate}
	\item We propose HyperMODEST, a universal method of filtering out pseudo-labels used for data augmentation with low confidence scores to accelerate self-training in MODEST \cite{modest}, a state-of-the-art unsupervised 3D LiDAR-based object detection pipeline.
	\item We validate our method on the nuScenes and Lyft datasets. On the nuScenes dataset, we demonstrate a significant improvement of 1.6\% in AP\_{BEV} in 0-80m range at IoU=0.25 and an improvement of 1.7\% in AP\_{BEV} in 0-80m range at IoU=0.5 after two rounds of training in comparison to ten rounds of training in the baseline approach. On the Lyft dataset, we demonstrate an improvement of 0.3\% in AP\_{BEV} in 0-80m range at IoU=0.25.
	\item We compare our method with score filtering on the pseudo-labels and score filtering on the pseudo-labels with static label retention and demonstrate the importance of preserving high recall during early rounds of self-training. 
\end{enumerate}

\section{RELATED WORK}

\noindent \textbf{Deep learning on 3D data:}  Existing 3D object detectors either take 3D point clouds generated by LiDAR as input or take image data and combine it with depth estimation. 3D LiDAR-based object detectors can be categorized according to their representation of the point cloud. One category of detectors operates on raw point clouds directly \cite{pointrcnn} \cite{voting-based(2021)}\cite{group-free(2021)}. These architectures improve the extracted feature quality by considering the local structure in point clouds. Other voxel-based methods include \cite{pdv(2022)} \cite{pv-rcnn++(2022)}\cite{centerpoint(2021)} \cite{infofocus}. These methods need to account for intensive density and intensity variation in the point cloud. Other recent architectures use multi-view representation, such as converting the 3D point cloud into BEV images, mesh representations or feature-based representations \cite{MVF-hybrid(2020)}\cite{AFDet(2020)} \cite{avod}.
\\

\noindent \textbf{Recognition with minimal labels:}
To generate pseudo-labels for 3D object detection, common heuristics include using surface normals \cite{surface_normal} and motion \cite{motion1}\cite{motion2} to distinguish foreground from background objects. Recent works aggregate unlabelled LiDAR information over the past traversals of the same route to produce pseudo-labels \cite{modest}\cite{hindsight}. In \cite{dream}, previously recorded traversals are replayed, and object tracks are smoothed forward and backward to interpolate detections. 

Multi-modal unsupervised learning and self-supervised pre-training with synchronized cameras and LiDAR data are also used for object detection and semantic segmentation \cite{drive_segment}\cite{pre-training}. In the absence of annotated data in the target domain, domain adaptation adapts a model to a new geographical location, different weather conditions, sensor modalities and from simulated data to real data. Different approaches include domain-invariant feature learning \cite{sf-uda}, cross-modal \cite{xmuda}, and semi-supervised \cite{co-training} domain adaptation. 
\\

\noindent \textbf{Self-training and curriculum learning:} Recent advances in domain adaptation show that self-training is highly effective for unsupervised domain adaptation \cite{curriculum_2} \cite{st3d}. Self-training is an iterative process of predicting detections and then taking the confident predictions as pseudo-labels for retraining. Self-training can also be regarded as noisy label learning \cite{noisy_2}. With noisy pseudo-labels, self-training can lead to error propagation and concept drift. To address this problem, curriculum learning provides an improvement over the standard training approach based on random data shuffling by training the machine learning model from easy to hard samples. \cite{curriculum_1}\cite{curriculum_2} propose a confidence-regularized and class-balanced self-training framework. Label and model regularization have been applied to image classification and semantic segmentation problems in the presence of annotated data from the source domain. Integrating curriculum learning in MODEST~\cite{modest} is challenging since the model is class-agnostic and there isn't a source domain. 
\\

\noindent \textbf{Network regularization: } Data augmentation is a widely used technique to increase the number of ground truths per point cloud and simulate objects existing in different environments \cite{second}. Other network regularization techniques include weight decay, label smoothing \cite{DL}, knowledge distillation and network output regularization \cite{DL}. 

\section{METHODOLOGY}
Since our method is based on MODEST~\cite{modest}, we will first briefly summarize its architecture. Then, we will describe three methods of score filtering that we experimented with and the motivations of our choice. The three methods are score filtering for pseudo-labels, score filtering for pseudo-labels with static labels retention and score filtering for data augmentation.

\subsection{Preliminaries}
MODEST \cite{modest} is a self-supervised 3D object detection pipeline that focuses on the detection of mobile objects. It includes two stages: seed label generation and self-training. In the first stage, for multiple traversals of the same scene, an ephemerality score (PP persistent point score) that captures the change of the LiDAR point's local neighbourhood across traversals is calculated for each LiDAR point. The PP score reflects the likelihood of a point being part of the static background. A high PP score implies that the point is more likely to be a part of the static background. Since a mobile object observed at a specific location in one traversal is unlikely to be observed at the same location in another traversal, it is an ephemeral member of a scene. Then, the LiDAR points are clustered with DBSCAN \cite{dbscan} clustering algorithm according to the ephemerality scores and their 3D coordinates. Clusters with high ephemerality scores represent mobile objects and are retained as seed-bounding boxes. The pipeline is class-agnostic. All objects have the common label ``Dynamic". 

In the second stage, a 3D object detector is trained iteratively. In the first round, the model is trained from scratch on the seed labels. The trained model is used to produce an improved set of pseudo ground-truth labels for the second round of training. In the second round, the detector is then trained from scratch with these newly generated intermediate pseudo-labels, and this process iterates. 

\subsection{Score filtering for pseudo-labels}
\begin{algorithm}
\caption{Filter-Pseudo-Labels}\label{algorithm1}
\begin{algorithmic}[1]
\Require \{$P_i$ \} LiDARs with accurate localization
\Require $I_{max}$ maximum self\_training (ST) iterations

\State $B_0$ $\leftarrow$ \{seed\_labels($P_i$)\}
\State $DB_0$ $\leftarrow$ $B_0$
\State $D_0$ $\leftarrow$ train\_detector(\{$P_i$\}, $B_0$)  
\For{$ j \leftarrow 1$ to $I_{max}$ }
    \State $B_j \leftarrow$ get\_detection($D_{j-1}$, \{$P_i$\}) 
    \State \tikzmark{start1}$ \bf{t} \leftarrow \bm{\rho}_.25(B_j)$\tikzmark{end1}
    \State $B_j \leftarrow$ filter\_by\_PP($B_j$) 
    \State \tikzmark{start2} $B_j \leftarrow$ \textbf{filter\_by\_confidence\_score\_t}($B_j$)    
    \State $DB_j$ $\leftarrow$ $\bf{B_j}$ \tikzmark{end2}
    \State $D_j \leftarrow$ train\_detector(\{$P_i$\}, $B_j$, $DB_j$) 
\EndFor \\
\Return $D_{I_{max}}$  
\end{algorithmic}
\end{algorithm}
\TextboxSmall{start1}{end1}{}
\Textbox{start2}{end2}{}

We performed a qualitative analysis of the original implementation in MODEST \cite{modest} on the nuScenes dataset. As shown in Fig. \ref{fig:nuscenes_result}, with each round of training, the model predicts more objects, but more false positives are produced. Although the seed labels are generated conservatively to favour high precision, they contain noise and only capture a subset of objects. During self-training, since the $i$'th round of training uses the pseudo-labels produced by the $i-1$'th round, new objects are discovered based on pseudo-labels that represent a smaller subset of all objects. As a result, noise can be amplified and carried to future training rounds. 

We used the idea of curriculum learning to mitigate this problem of concept drift and to help the model learn from easier to more difficult examples. In each round of training, we filter out 25\% of intermediate pseudo-labels with low confidence scores using the threshold $\rho_.25(B_j)$, as outlined in Algorithm \ref{algorithm1}. $B_j$ is the set of pseudo-labels, $DB_j$ is the pseudo ground-truth database used for data augmentation. Our modifications of the original algorithm presented in MODEST\cite{modest} are in bold. 

\subsection{Score filtering for pseudo-labels with static labels retention}
Similarly to MODEST~\cite{modest}, we observed that the detector can identify more static objects such as parked cars through self-training. Since these parked cars appear across multiple traversals, they were not captured in the seed labels generation. However, since they are similar in shape to the mobile objects in seed labels, they are discovered during self-training. In MODEST~\cite{modest}, pseudo-labels are removed if the $\alpha$ percentile of the PP scores within the box is larger than $\gamma$. A high PP score implies that the point is more likely to be a part of the static background. In Algorithm \ref{algorithm2}, on top of the score filtering described in section B, in the function filter\_by\_PP\_And\_Keep\_Static, we keep pseudo-labels with confidence scores higher than a threshold of 0.8 among those that are to be removed due to their PP scores. We wish to retain more correct predictions of static objects.

\begin{algorithm}
\caption{Filter-And-Keep-Static}\label{algorithm2}
\begin{algorithmic}[1]
\Require \{$P_i$ \} LiDARs with accurate localization
\Require $I_{max}$ maximum self\_training (ST) iterations

\State $B_0$ $\leftarrow$ \{seed\_labels($P_i$)\}
\State $DB_0$ $\leftarrow$ $B_0$
\State $D_0$ $\leftarrow$ train\_detector(\{$P_i$\}, $B_0$)  
\For{$ j \leftarrow 1$ to $I_{max}$ }
    \State $B_j \leftarrow$ get\_detection($D_{j-1}$, \{$P_i$\}) 
    \State \tikzmark{start3} $ \bf{t} \leftarrow \bm{\rho}_.25(B_j)$
    \State $B_j \leftarrow$ \textbf{filter\_by\_PP\_And\_Keep\_Static}($B_j$) 
    \State $B_j \leftarrow$ \textbf{filter\_by\_confidence\_score\_t}($B_j$)    
    \State $DB_j$ $\leftarrow$ $\bf{B_j}$ \tikzmark{end3}
    \State $D_j \leftarrow$ train\_detector(\{$P_i$\}, $B_j$, $DB_j$) 
\EndFor \\
\Return $D_{I_{max}}$  
\end{algorithmic}
\end{algorithm}
\Textbox{start3}{end3}{}

\begin{table*}[!t]
	\renewcommand{\arraystretch}{1.3}
	\centering
       
	\begin{tabular}{|c|c| c|c|c|c|m {5em}|}
		\hline
		Dataset & Round & Method & Pseudo-labels & \# Pseudo-labels & Pseudo gt-database & \# in pseudo gt-database \\
		\hline
		nuScenes & 1 & Filter-Data-Augmentation & NA & 8,357 & 20\%, score\_threshold=0.20 & 6,611 \\
		\hline
		nuScenes & 2& Filter-Data-Augmentation & NA & 15,671 & 20\%, score\_threshold=0.22 & 12,537 \\
		\hline
		nuScenes & 3 & Filter-Data-Augmentation & NA & 24,650 & 20\%, score\_threshold=0.31 & 19,720 \\
		\hline
		nuScenes & 4 & Filter-Data-Augmentation & NA & 31,251 & 20\%, score\_threshold=0.49 & 24,999 \\
		\hline
		nuScenes & 1 & Filter-Pseudo-Labels & 25\%,  score\_threshold=0.21& 6,536 & NA & 6,536 \\
		\hline
		nuScenes & 1 & Filter-And-Keep-Static & 25\%,  score\_threshold=0.21& 6,603 & NA & 6,603 \\
		\hline
		Lyft & 1 & Filter-Data-Augmentation & NA & 69,880 & 20\%, score\_threshold=0.56 & 55,899 \\
		\hline
		Lyft & 1 & Filter-Pseudo-Labels& 10\%, score\_threshold=0.27 & 64,560 & NA & 64,560 \\
		\hline
		Lyft & 1 & Filter-And-Keep-Static& 10\%, score\_threshold=0.27  & 66,249 & NA & 66,249 \\
		\hline
		
	\end{tabular}
  \caption{Filtering methods used for experiments on nuScenes and Lyft}
	
\label{Table:Score_filtering}
\end{table*}

\subsection{HyperMODEST: Score filtering for data augmentation}
Filter-Pseudo-Labels and Filter-And-Keep-Static are strategies that improve the model's performance, as shown in section V. However, in our unsupervised training setting, tuning the score threshold with which we filter out intermediate pseudo-labels is challenging. Since we don't have access to the ground truth labels of the training set, with the above processing techniques of pseudo-labels, we make assumptions on the accuracy of the shape and localization of the detections. If the score threshold is too high, we might risk losing true positives. The percentile of the score threshold also needs to adapt to different datasets.

We propose HyperMODEST, a universal method that can largely accelerate the self-training process and is independent of the data domain. First, we follow MODEST~\cite{modest} to create the pseudo-labels based on the detection output from the previous round of training. Second, to generate the pseudo ground-truth database for data augmentation, we filter out 20\% of intermediate pseudo-labels with low scores. The method is described in Algorithm \ref{algorithm3}. 

\begin{algorithm}
\caption{Filter-Data-Augmentation}\label{algorithm3}
\begin{algorithmic}[1]
\Require \{$P_i$ \} LiDARs with accurate localization
\Require $I_{max}$ maximum self\_training (ST) iterations

\State $B_0$ $\leftarrow$ \{seed\_labels($P_i$)\}
\State $DB_0$ $\leftarrow$ $B_0$
\State $D_0$ $\leftarrow$ train\_detector(\{$P_i$\}, $B_0$)  
\For{$ j \leftarrow 1$ to $I_{max}$ }
    \State $B_j \leftarrow$ get\_detection($D_{j-1}$, \{$P_i$\}) 
    \State \tikzmark{start4}$ \bf{t} \leftarrow \bm{\rho}_.25(B_j)$\tikzmark{end4}
    \State $B_j \leftarrow$ filter\_by\_PP($B_j$)  
    \State \tikzmark{start5} $DB_j \leftarrow$ \textbf{filter\_by\_confidence\_score\_t}($B_j$) \tikzmark{end5}
    \State $D_j \leftarrow$ train\_detector(\{$P_i$\}, $B_j$, $DB_j$) 
\EndFor \\
\Return $D_{I_{max}}$  
\end{algorithmic}
\end{algorithm}
\TextboxSmall{start4}{end4}{}
\TextboxSmall{start5}{end5}{}

\begin{table*}[!t]
	\renewcommand{\arraystretch}{1.3}
	\centering
       
	\begin{tabular}{|m {9em}||c|c| c|c|c|c|c|c|m {3em}|}
		\hline
		\multirow{2}{4em}{Method} & \multicolumn{4}{c|}{$AP_{BEV}/AP_{3D}$ at IoU=0.25} &
		\multicolumn{4}{c|}{$AP_{BEV}/AP_{3D}$ at IoU=0.5} & \# predicted objects\\
		\cline{2-9}
		& 0-30 & 30-50 & 50-80 & 0-80 &0-30 & 30-50 & 50-80 & 0-80 & \\
		\hline
 MODEST, R1 & 24.8 / 18.7  & 2.8 / 1.5 & 1.1 / 0.1 & 10.5 / 7.5 & 12.9 / 5.0 & 0.9 / 0.0 & 0.1 / 0.0 & 4.8 / 1.4 &  6.8\\
  \hline 
		\textbf{MODEST, R10} \cite{modest} & \textbf{24.8 / 17.1} & \textbf{5.5 / 1.4} & \textbf{1.5 / 0.3} & \textbf{11.8 / 6.6} & \textbf{11.0 /  7.6} & \textbf{0.4 / 0.0} & \textbf{0.0 / 0.0} & \textbf{3.9 / 2.2} & -  \\
		\hline
		HyperMODEST, R1 & 26.5 / 19.7  & 3.5 / 1.7 & 1.3 / 0.2 & 12.1 / 8.0 & 13.5 / 5.3 & 0.9 / 0.0 & 0.1 / 0.0 & 5.4 / 1.6 &  6.12\\
		\hline
		\textbf{HyperMODEST, R2} & \textbf{28.1 / 20.5} & \textbf{4.8 / 2.3} & \textbf{1.9 / 0.4} & \textbf{13.4 / 8.7} & \textbf{14.2 / 5.7} & \textbf{1.4 / 0.0} & \textbf{0.2 / 0.0} & \textbf{5.6 / 1.6} & 9.02 \\ 
		\hline
		HyperMODEST, R3 &27.6 / 20.4 & 5.2 / 2.2 & 2.4 / 0.4 & 13.1 / 8.5 &  13.8 / 6.3 & 1.2 / 0.0 & 0.3 / 0.0 & 5.9 / 1.9& 11.18 \\ 
		\hline
		HyperMODEST, R4 & 26.6 / 19.7 & 4.7 / 1.9 & 2.5 / 0.5 & 12.3 / 7.9 & 13.6 / 7.0 & 1.0 / 0.0 & 0.2 / 0.0 & 5.4 / 2.1 & 12.4 \\
		\hline
		\color{red}\textbf{Improvement} & \color{red}\textbf{+3.3 / + 3.4}    & \color{red}\textbf{-0.7 / +0.9 }  & \color{red}\textbf{ +0.4 / +0.1}  & \color{red}\textbf{+1.6 / +2.1}  & \color{red}\textbf{+3.2 / -1.9} &\color{red}\textbf{ +1.0 / 0.0} & \color{red}\textbf{+0.2 / 0.0}  & \color{red}\textbf{+1.7 / -0.6 } &-\\
		\hline
		Supervised \cite{modest} & 39.8 / 34.5 & 12.9 / 10.0 & 4.4 / 2.9 & 22.2 / 18.2 & 29.5 / 26.3 & 8.4 / 6.1 & 2.4 / 1.1 & 15.5 / 13.3 & - \\
		\hline
		
	\end{tabular}
 \caption{ 3D detection results with Filter-Data-Augmentation on nuScenes Dataset. The results are compared with the performance of Round 10 (R10) in the baseline from MODEST \cite{modest}. Improvement is shown between ``HyperMODEST, R2" and ``MODEST, R10".  }
	\label{Table: nuscenes}
\end{table*}

\section{EXPERIMENTAL SETUP}
In this section, we introduce the two datasets used in our experiments, the training setup, the score filtering strategies and evaluation metrics.

\subsection{Datasets}
The nuScenes \cite{nuscenes} dataset is an autonomous driving dataset for 3D perception tasks. It contains 1,000 driving sequences, each 20 seconds long. LiDAR scans are collected by a 32-beam roof LiDAR. The Lyft \cite{lyft} dataset contains more than 350 scenes at 60-90 minutes long. LiDAR scans are collected by a 64-beam roof LiDAR. The data are pre-processed as in MODEST~\cite{modest} to filter out locations with less than 2 examples in the training set, which results in a train/test split of 3,985/2,324 keyframes for nuScenes and 11,873/4,901 point clouds for Lyft. Both datasets are converted into KITTI dataset format for the downstream detection task. All the annotated ground-truth objects are assigned a common “Dynamic” label. The ground-truth labels are only used for evaluation, not training. 

\subsection{Implementation details}
We use the parameters and implementation from MODEST \cite{modest} to generate the seed labels. Our training pipeline is based on the published code of MODEST \cite{modest} and the open-sourced OpenPCDet \cite{openpcdet}. We present results on PointRCNN \cite{pointrcnn}. For nuScenes dataset, the model is trained for 4 rounds with the method Filter-Data-Augmentation, 80 epochs each with a batch size of 4. It is trained for 1 round with Filter-Pseudo-Labels and Filter-And-Keep-Static. The perception range is up to 70m. For Lyft dataset, the model is trained for 1 round for each of the three methods, 60 epochs with a batch size of 2. The perception range is up to 90m. We use the default hyperparameters in MODEST \cite{modest} and OpenPCDet~\cite{openpcdet} for the model with binary cross-entropy loss. The Adam optimizer is used with an initial learning rate of 0.01. Data augmentations including random flipping, vertical rotation, global scaling and ground-truth sampling are used to boost the performance of the 3D object detectors. For both datasets, each sample contains 40 labels after ground-truth sampling. The models are trained with two NVIDIA GeForce 2080 GPUs. 

\subsection{Score filtering scheme}
In Table \ref{Table:Score_filtering}, we present the score filtering methods used for each experiment. The column ``Method" refers to algorithms described in section III. The column ``Pseudo-labels" refers to the percentile used to filter out labels. Note that with method Filter-And-Keep-Static, a high\_threshold of 0.8 is used to retain static objects as described in Algorithm 2. For the Lyft dataset, a lower percentile is used to lower the score threshold at the beginning of the self-training. The column ``Pseudo-gt-database" refers to the percentile used to filter out labels in the pseudo ground-truth database solely used for data augmentation in Filter-Data-Augmentation.  

\subsection{Evaluation metrics}
We follow MODEST \cite{modest} to evaluate object detection in the bird’s-eye view (BEV) and in 3D at various depth ranges for mobile objects. The classification is binary. We present average precision (AP) with the intersection over union (IoU) thresholds at 0.5/0.25. 

\begin{table*}[!t]
	\renewcommand{\arraystretch}{1.3}
	\centering
         
	\begin{tabular}{|m {13em}||c|c| c|c|c|c|c|c|m {4em}|}
		\hline
		\multirow{2}{4em}{Method} & \multicolumn{4}{c|}{$AP_{BEV}/AP_{3D}$ at IoU=0.25} &
		\multicolumn{4}{c|}{$AP_{BEV}/AP_{3D}$ at IoU=0.5} & Number of predicted objects\\
		\cline{2-9}
		& 0-30 & 30-50 & 50-80 & 0-80 &0-30 & 30-50 & 50-80 & 0-80 & \\
		\hline
		Baseline MODEST, R1 & 24.8 / 18.7 & 2.8 / 1.5 &  1.1 / 0.1 & 10.5 / 7.5  &12.9 / 5.0 & 0.9 / 0.0 & \color{red}\textbf{0.1} \color{black}/ 0.0 & 4.8 / 1.4 & 6.8\\
		\hline
		\textbf{Filter-Data-Augmentation, R1} & \color{red}\textbf{26.5} \color{black}/ 19.7  & \color{red}\textbf{3.5} \color{black}/ 1.7 & \color{red}\textbf{1.3 / 0.2} & \color{red}\textbf{12.1} \color{black}/ 8.0 & \color{red}\textbf{13.5} / \color{black}5.3 & 0.9 / 0.0 & \color{red}\textbf{0.1} \color{black}/ 0.0 & \color{red}\textbf{5.4 / 1.6} &  6.12\\
		\hline
		Filter-Pseudo-Labels, R1 & 25.6 / \color{red}\textbf{20.0} & \color{red}\textbf{3.5} / \color{red}\textbf{1.9} & \color{red}\textbf{1.3 / 0.2} & 11.4 / \color{red}\textbf{8.1} & \color{red}\textbf{13.5} / \color{black}5.3 & \color{red}\textbf{1.1} \color{black}/ 0.0 & \color{red}\textbf{0.1} \color{black}/ 0.0 & 5.0 / 1.6 & 5.66 \\
		\hline
		Filter-And-Keep-Static, R1 & 25.3 / 19.4 & 3.1 / 1.4 & 1.0 / \color{red}\textbf{0.2} & 11.3 / 7.8 & 12.1 / \color{red}\textbf{8.0} & 1.0 / 0.1 & \color{red}\textbf{0.1} \color{black}/ 0.0 & 4.9 / 1.6 & 5.40 \\
		\hline
	\end{tabular}
\caption{Comparison of Filter-Pseudo-Labels, Filter-And-Keep-Static and Filter-Data-Augmentation on nuScenes Dataset. R1 stands for Round 1. }
\label{Table: nuscenes2}
\end{table*}

\begin{figure*}[!t]
	\centering 
	\includegraphics[width=0.9 \textwidth]{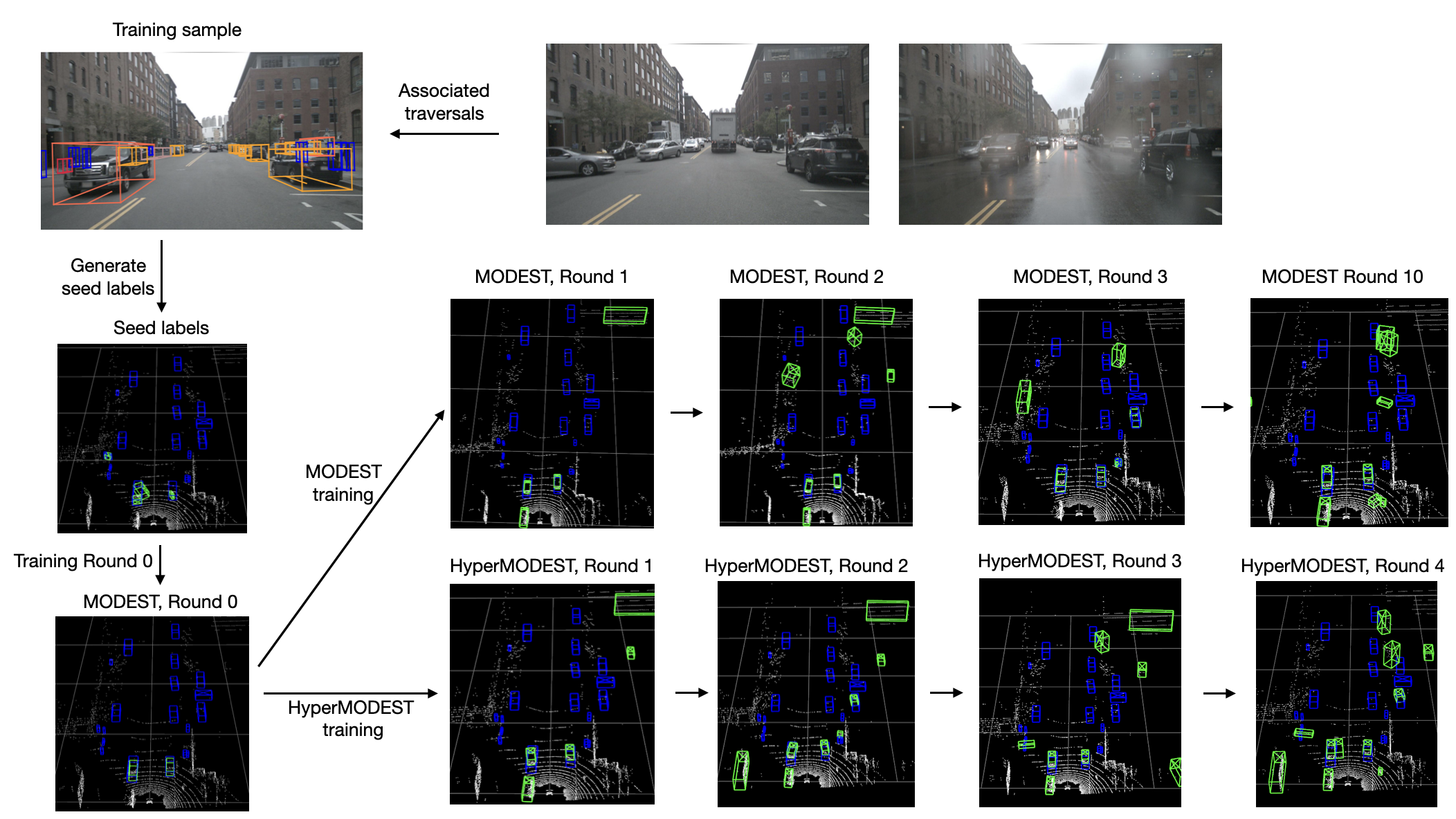} 
	\caption{Comparison between HyperMODEST and MODEST \cite{modest} after each round of training on nuScenes dataset. The ground truth annotations are shown in blue, the detections are shown in green.
		1), Seed labels generation: the seed labels for mobile objects are generated from the training sample by leveraging associated traversals of the same location. 2), Round 0 training: the pseudo ground-truth labels for training are the generated seed labels. 3), ``MODEST" Round 1,2,3 and 10 show the model output with the baseline implementation in MODEST \cite{modest}. 4), ``HyperMODEST" Round 1,2,3 and 4 show the model output with the addition of Filter-Data-Augmentation.}
	\label{fig:nuscenes_result}
\end{figure*}

\section{RESULTS}
In this section, we first compare the performance of HyperMODEST with MODEST on the nuScenes dataset. Second, we compare Filter-Data-Augmentation with Filter-And-Keep-Static and Filter-Pseudo-Labels with one round of iterative self-training on nuScenes and Lyft dataset. Then, we perform a qualitative analysis for HyperMODEST. Lastly, we evaluate the quality of pseudo-labels in the training set against the ground-truth.

\subsection{ Filtering for data augmentation leads to faster self-training}

Table \ref{Table: nuscenes} shows a comparison between the baseline and Filter-Data-Augmentation. First, by comparing ``HyperMODEST, R1" with ``MODEST, R10", we observe that the addition of filtering for data augmentation helps the model to achieve higher performance after only one round of training. Second, by comparing ``HyperMODEST, R2" with ``MODEST, R10", we observe a significant improvement of 3.3\% in AP\_{BEV} at IoU=0.25 in 0-30m range and an improvement of 1.6\% in AP\_{BEV} at IoU=0.25 in 0-80m range. Third, although the addition of pseudo ground-truth database filtering helps the model to outperform the baseline, the performance does not improve beyond Round 3. In fact, AP\_{BEV} at IoU=0.25 in 0-80m range is 12.3\% for ``HyperMODEST, R4", lower than 13.4\% for ``HyperMODEST, R2".  This shows the limitation of the self-training algorithm because the model is trained iteratively on a set of noisy seed labels. The baseline ``MODEST, R1, R10" also shows a decrease in performance in certain metrics. By comparing ``HyperMODEST, R2" with ``Supervised", we note that there are still significant improvements to be made for unsupervised training to match the performance of supervised training. 

\begin{table*}[!t]
	\renewcommand{\arraystretch}{1.3}
	\centering
       
	\begin{tabular}{|m {13em}||c|c| c|c|c|c|c|c|m {3em}|}
		\hline
		\multirow{2}{4em}{Method} & \multicolumn{4}{c|}{$AP_{BEV}/AP_{3D}$ at IoU=0.25} &
		\multicolumn{4}{c|}{$AP_{BEV}/AP_{3D}$ at IoU=0.5}\\
		\cline{2-9}
		& 0-30 & 30-50 & 50-80 & 0-80 &0-30 & 30-50 & 50-80 & 0-80 \\
		\hline
		\textbf{Baseline MODEST, R1} & \textbf{69.2 / 66.8} &  \textbf{49.3 / 44.9} & \textbf{11.5 / 9.1} & \textbf{46.1 / 43.2} &  \textbf{51.0 / 46.0} & \textbf{34.3 / 23.0} & \textbf{5.3 / 2.2} & \textbf{31.7 / 24.9}\\
		\hline
		\textbf{HyperMODEST, R1} & \textbf{69.6 / 67.1} & \textbf{48.5 / 45.4} & \textbf{11.9 / 9.3} & \textbf{46.4 / 43.7} & \textbf{50.5 / 45.6} & \textbf{34.8 / 23.8} & \textbf{5.4 / 1.9} & \textbf{31.6 / 25.0} \\
		\hline
            \color{red}\textbf{Improvement} & \color{red}\textbf{+0.4 / +0.3} & \color{red}\textbf{-0.8 / +0.5} & \color{red}\textbf{+0.4 / +0.2} & \color{red}\textbf{ +0.3 / +0.5} & \color{red}\textbf{-0.5 / -0.4} & \color{red}\textbf{+0.5 / +0.8} & \color{red}\textbf{+0.1 / -0.3} & \color{red}\textbf{-0.1 / +0.1} \\
            \hline
		Filter-Pseudo-Labels, R1 &  69.3 / 66.8 & 48.4 / 44.4 & 11.5 / 8.6 & 46.2 / 43.3 & 50.6 / 45.6 & 33.9 / 23.6 & 4.9 / 2.3 & 31.5 / 25.0 \\
		\hline
		Filter-And-Keep-Static, R1 &69.3 / 66.8 & 48.0 / 44.2 & 11.7 / 9.1 &46.0 / 43.2 &   51.1 / 46.2 & 34.0 / 23.2 & 5.4 / 2.0 & 31.7 / 25.0 \\
        \hline
		Supervised \cite{modest} & 82.8 / 82.6 & 70.8 / 70.3 & 50.2 / 49.6 & 69.5 / 69.1 & 72.3 / 69.5 & 53.2 / 48.1 & 27.9 / 20.5 & 53.1 / 48.1  \\
		\hline
	\end{tabular}
   \caption{Comparison of Filter-Pseudo-Labels, Filter-And-Keep-Static and Filter-Data-Augmentation (HyperMODEST) on Lyft Dataset. R1 stands for Round 1. Improvement is shown between ``Baseline MODEST, R1" and ``HyperMODEST, R1".}
	\label{Table: lyft}
\end{table*}

\begin{table*}[!t]
	\renewcommand{\arraystretch}{1.3}
	\centering
         
	\begin{tabular}{|m {19em}||c|c| c|c|c|c|c|c|}
		\hline
		\multirow{2}{4em}{Method} & \multicolumn{4}{c|}{$AP_{BEV}/AP_{3D}$ at IoU=0.25} &
		\multicolumn{4}{c|}{$AP_{BEV}/AP_{3D}$ at IoU=0.5} \\
		\cline{2-9}
		& 0-30 & 30-50 & 50-80 & 0-80 &0-30 & 30-50 & 50-80 & 0-80  \\
		\hline
		Round 0 output &\color{red}\textbf{22.7 / 19.9}  & 1.6 / \color{red}\textbf{0.7} & 0.2 / 0.1 & 9.1 / \color{red}\textbf{8.1} & \color{red}\textbf{13.6 / 6.4} & 0.3 / 0.0 & 0.0 / 0.0 & \color{red}\textbf{5.3} \color{black}/ 1.7 \\
		\hline 
		1. Filter-Pseudo-Labels, pseudo-labels & 19.2 / 18.5 & \color{red}\textbf{2.0} \color{black}/ 0.6 & \color{red}\textbf{0.4 / 0.2} & \color{red}\textbf{9.4} \color{black}/ 7.1 & 12.4 / \color{red}\textbf{6.4} & \color{red}\textbf{0.4 / 0.1} & \color{red}\textbf{0.1} \color{black}/ 0.0 & 4.4 / \color{red}\textbf{1.9} \\
		\hline 
		1. Filter-Pseudo-Labels, pseudo gt-database & 19.2 / 18.5 & \color{red}\textbf{2.0} \color{black}/ 0.6 & \color{red}\textbf{0.4 / 0.2} & \color{red}\textbf{9.4} \color{black}/ 7.1 & 12.4 / \color{red}\textbf{6.4} & \color{red}\textbf{0.4 / 0.1} & \color{red}\textbf{0.1} \color{black}/ 0.0 & 4.4 / \color{red}\textbf{1.9} \\
		\hline
		2. Filter-And-Keep-Static, pseudo-labels &19.1 / 18.5 & \color{red}\textbf{2.0} \color{black}/ 0.6 & \color{red}\textbf{0.4 / 0.2} & \color{red}\textbf{9.4} \color{black}/ 7.1 & 12.4 / \color{red}\textbf{6.4} & \color{red}\textbf{0.4 / 0.1} & \color{red}\textbf{0.1} \color{black}/ 0.0 & 4.4 / \color{red}\textbf{1.9} \\
		\hline 
		2. Filter-And-Keep-Static, pseudo gt-database &19.1 / 18.5 & \color{red}\textbf{2.0} \color{black}/ 0.6 & \color{red}\textbf{0.4 / 0.2} & \color{red}\textbf{9.4} \color{black}/ 7.1 & 12.4 / \color{red}\textbf{6.4} & \color{red}\textbf{0.4 / 0.1} & \color{red}\textbf{0.1} \color{black}/ 0.0 & 4.4 / \color{red}\textbf{1.9} \\
		\hline
		\textbf{3. Filter-Data-Augmentation, pseudo-labels} & 20.9 / 18.3 & 1.9 / 0.4 & 0.3 / 0.1 & 9.2 / 6.9 & 12.2 / 6.2 & 0.3 / 0.0 & 0.0 / 0.0 & 4.2 / 1.8\\
		\hline
		\textbf{3. Filter-Data-Augmentation, pseudo gt-database} & 19.2 / 18.5 & \color{red}\textbf{2.0} \color{black}/ 0.6 & \color{red}\textbf{0.4 / 0.2 }& \color{red}\textbf{9.4} \color{black}/ 7.1 & 12.4 / 6.3 & \color{red}\textbf{0.4 / 0.1} &\color{red}\textbf{0.1} \color{black}/ 0.0 &  4.4 / \color{red}\textbf{1.9} \\
		\hline
	\end{tabular}

 \caption{Quality of pseudo-labels and labels for data augmentation in pseudo ground-truth database for Round 1 training with the 3 methods on nuScenes dataset. The pseudo-labels are evaluated against the ground truth. ``Round 0" output indicates the evaluation of the model after round 0 training on the training set against the ground truth. }
	\label{Table: nuscenes_labels}
\end{table*}

\subsection{Filtering for data augmentation outperforms pseudo-labels filtering}
Table \ref{Table: nuscenes2} shows a comparison between the three filtering algorithms for Round 1 training on the nuScenes dataset. First, all three algorithms outperform the baseline, indicating that processing intermediate pseudo-labels by filtering out low confidence score labels helps the model to discover more objects with higher precision. Second, Filter-Pseudo-Labels performs slightly better than Filter-And-Keep-Static, indicating that removing pseudo-labels if the $\alpha$ percentile of the PP scores within the box is larger than $\gamma$ is effective in removing false positives. Third, Filter-Data-Augmentation outperforms Filter-Pseudo-Labels and Filter-And-Keep-Static. Since Filter-Pseudo-Labels and Filter-And-Keep-Static apply score filtering on the pseudo-labels, they might have filtered out true positives. In contrast, Filter-Data-Augmentation applies score filtering only on the pseudo ground-truth database for data augmentation. In other words, Filter-Data-Augmentation preserves all the pseudo-labels generated from the previous round of training.  It is more practical and more generalizable to a variety of datasets since it preserves all pseudo-labels while only using labels with high confidence scores for data augmentation. 

\subsection{Generalization to Lyft dataset}
Table \ref{Table: lyft} shows a comparison between the three filtering algorithms for Round 1 training on the Lyft dataset. Similar to the experiments on nuScenes, the method Filter-Data-Augmentation outperforms Filter-Pseudo-Labels and Filter-And-Keep-Static. Compared to the baseline, Filter-Data-Augmentation improves the AP\_{BEV} at IoU=0.25 by 0.3\% in 0-80m range and by 0.5\% for AP\_{3D}. Although the improvement is minor, it can be shown that our method can be generalized to another dataset with a different sensor modality. Lyft dataset \cite{lyft} is collected with a 64-beam LiDAR whereas nuScenes dataset \cite{nuscenes} is collected with a 32-beam LiDAR.

\subsection{Qualitative analysis}
Fig. \ref{fig:nuscenes_result} shows a comparison of model predictions between the baseline and Filter-Data-Augmentation on a training sample in the nuScenes dataset. First, by comparing the predictions of ``HyperMODEST, Round 2" with ``MODEST, Round 10", ``HyperMODEST, Round 2" predicts correctly some of the static objects on the right side of the scene where ``MODEST, Round 10" fails to predict. Second, by comparing the predictions of ``HyperMODEST, Round 2" with ``MODEST, Round 2", ``HyperMODEST, Round 2" has higher recall since it predicts more static objects. Our qualitative analysis shows the benefit of score filtering for data augmentation, and is consistent with the previous quantitative analysis. We should note that in ``HyperMODEST Round 4", we observe more false positives and a loss of true positives detected in ``HyperMODEST Round 2". The noise in the seed labels is amplified through self-training, and score filtering for data augmentation alone is not enough to regularize the model although it allows the model to learn faster. It is worth exploring other regularization techniques to help the model learn on unlabeled data progressively.

\subsection{Ablation studies}

\noindent \textbf{Quality of training labels:}
To analyze the effect of score filtering on the performance of the first round of training on nuScenes dataset, Table \ref{Table: nuscenes_labels} shows the AP\_{BEV} and AP\_{3D} at IoU=0.25 and IoU=0.5 for the pseudo-labels and the labels in the pseudo ground-truth database solely used for data augmentation, for each of the 3 methods. The labels are evaluated against the ground-truth labels. By comparing ``Filter-Data-Augmentation, pseudo-labels" and ``Filter-Pseudo-Labels, pseudo-labels", Filter-Pseudo-Labels has higher AP\_{BEV} at IoU=0.25 and at IoU=0.5 in 0-80m range. Filter-Pseudo-Labels filters out 25\% of labels with low scores, whereas Filter-Data-Augmentation doesn't filter out any labels. The quality of the labels in the pseudo ground-truth database is similar for Filter-Pseudo-Labels and Filter-Data-Augmentation. Although the quality of the pseudo-labels is better for Filter-Pseudo-Labels and Filter-And-Keep-Labels, Table \ref{Table: nuscenes2} shows that their detection performance is poorer than Filter-Data-Augmentation. By filtering out pseudo-labels with low scores, we improve the precision while lowering the recall. This leads to an interesting insight. At the beginning of self-training, our pseudo-labels are only a small pool of ground truth labels. Thus, having higher recall for the pseudo-labels is essential for early rounds of training. 

\section{CONCLUSION}
In this work, we propose HyperMODEST, a universal method of filtering pseudo-labels with low confidence scores for data augmentation implemented on top of MODEST \cite{modest}, a state-of-the-art unsupervised 3D object detection pipeline. Through this simple strategy, we observe a consistent improvement on two large-scale datasets in comparison to the original implementation of MODEST \cite{modest} while using significantly less training time. We consider our work to offer important insights into regularization techniques' role in unsupervised training. The potential impact of accurate unsupervised object detection based on multiple traversals of the same location is monumental, allowing fast edge computation on self-driving cars, promoting fast deployment of self-driving technologies in any geographical location, and promoting the testing of new LiDAR sensor technologies.

\end{document}